\PassOptionsToPackage{main=english,russian}{babel}
\PassOptionsToPackage{T2A}{fontenc}
\documentclass[onecolumn]{misraj}
\DeclareUnicodeCharacter{02BE}{\textain}
\usepackage{lipsum}
\usepackage{pdfpages}
\usepackage{colortbl}
\usepackage{tabulary}
\usepackage{longtable}
\usepackage{array}
\usepackage{placeins}
\usepackage{tablefootnote}
\usepackage{tcolorbox}
\usepackage{wrapfig}
\usepackage{adjustbox}
\usepackage{float}
\usepackage{caption}
\usepackage{tabularx}
\usepackage{breqn} % For multiline equations
\usepackage{xspace}
\usepackage{pifont}
\definecolor{deeppurple}{HTML}{9e02f7}
\definecolor{forestgreen}{HTML}{2e7d43}
\usepackage{multirow}
\usepackage{tabularx}

\usepackage[utf8]{inputenc}

\usepackage{arydshln}

\usepackage{xspace} % 
\usepackage{makecell} % For line breaks in table cells
\usepackage{multirow}

\usepackage[framemethod=TikZ]{mdframed}
\usepackage{tcolorbox}
\usepackage{multicol}

\usepackage[utf8]{inputenc}
\usepackage[T1]{fontenc}
\usepackage[english]{babel} % don't use [arabic] here with arabtex
\usepackage{arabtex}
\usepackage{utf8}
\setcode{utf8}

\definecolor{lightblue}{RGB}{211, 227, 252} % Light blue => datacard
\definecolor{bgblue}{RGB}{247, 250, 255} % datacard background

\newcommand*\colourcheck[1]{%
  \expandafter\newcommand\csname #1check\endcsname{\textcolor{#1}{\ding{52}}}%
}
\newcommand*\colourcross[1]{%
  \expandafter\newcommand\csname #1cross\endcsname{\textcolor{#1}{\ding{55}}}%
}
\DeclareSymbolFont{extraup}{U}{zavm}{m}{n}
\DeclareMathSymbol{\vardiamond}{\mathalpha}{extraup}{87}
\colourcheck{blue}
\colourcheck{green}
\colourcheck{red}

\colourcross{blue}
\colourcross{green}
\colourcross{red}
\usepackage{nicematrix}
\usepackage{comment}
\usepackage{amssymb}
\usepackage{arabtex}
\usepackage[utf8]{inputenc}
\usepackage[bottom]{footmisc}
\usepackage{tipa,vowel}
\RequirePackage[hidelinks,colorlinks=true,linkcolor=blue,citecolor=blue]{hyperref}
\setcitestyle{number,square}
\title{\raisebox{-0.5em}{\includegraphics[width=2cm]{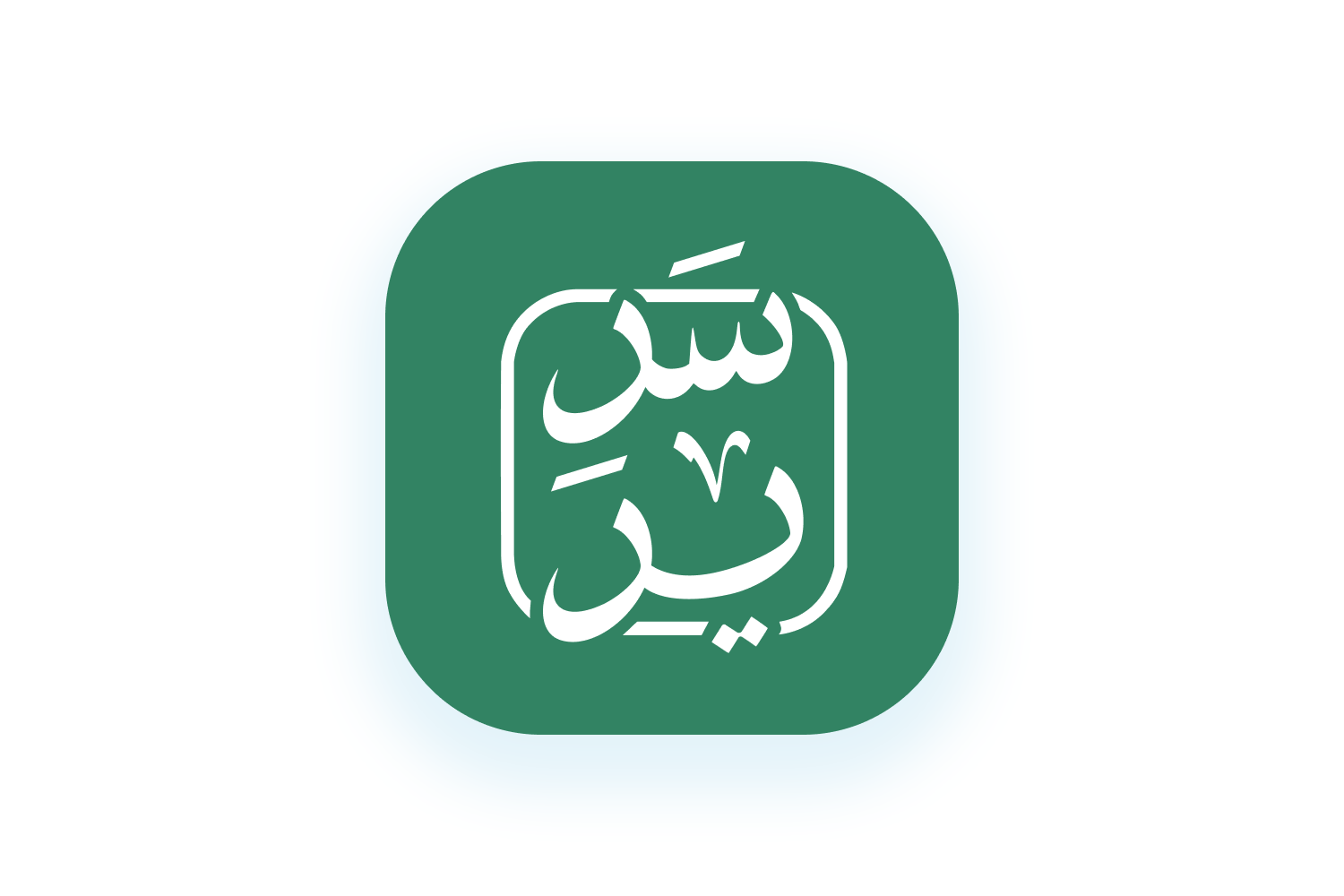}}Sadeed: Advancing Arabic Diacritization Through Small Language Model}

\multiauthors

\author{
    name={Zeina Aldallal},
    email={aldallal@misraj.ai}
}

\author{
    name={Sara Chrouf},
    email={sara.chrouf@misraj.ai}
}
\author{
    name={Khalil Hennara},
    email={hennara@misraj.ai}
}
\author{
    name={Mohamed Motasim Hamed},
    email={hamed@misraj.ai}
}

\author{
    name={Muhammad Hreden},  
    email={hreden@misraj.ai}
}
\author{
    name={Safwan AlModhayan},
    email={safwan@misraj.ai}
}

\affiliations{
     \includegraphics[width=2cm]{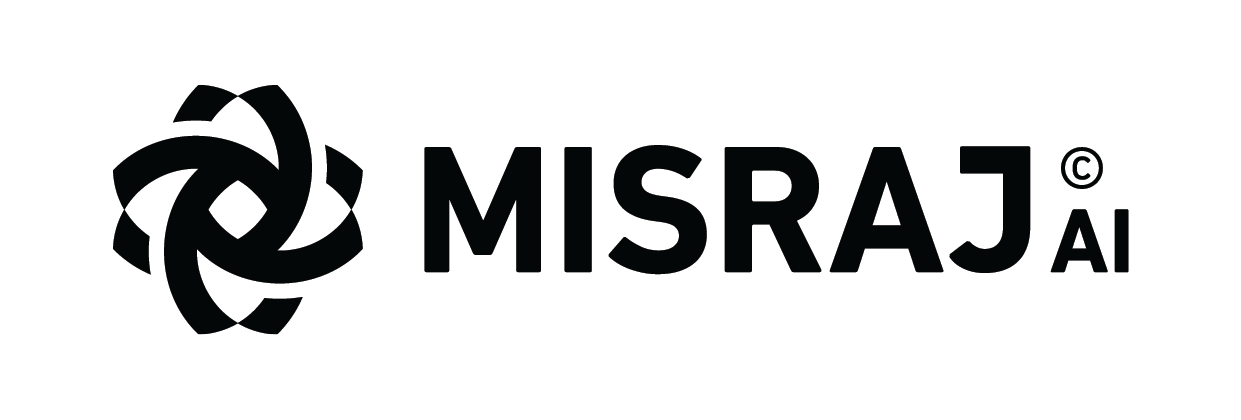}\\
     Khobar, Saudi Arabia \\
     {aldallal,chrouf,hennara,hamed,hreden,safwan}@misraj.ai \\
}
\date{\today}
\abstract{Arabic text diacritization remains a persistent challenge in natural language processing due to the language’s morphological richness. In this paper, we introduce \textbf{Sadeed}\textsuperscript{\ddag}, a novel approach based on a fine-tuned decoder-only language model adapted from \textbf{Kuwain 1.5B} \cite{hennara2025kuwain15barabicslm}, a compact model originally trained on diverse Arabic corpora. Sadeed is fine-tuned on carefully curated, high-quality diacritized datasets, constructed through a rigorous data-cleaning and normalization pipeline. Despite utilizing modest computational resources, Sadeed achieves competitive results compared to proprietary large language models and outperforms traditional models trained on similar domains. Additionally, we highlight key limitations in current benchmarking practices for Arabic diacritization. To address these issues, we introduce \textbf{SadeedDiac-25}, a new benchmark designed to enable fairer and more comprehensive evaluation across diverse text genres and complexity levels. Together, Sadeed and SadeedDiac-25 provide a robust foundation for advancing Arabic NLP applications, including machine translation, text-to-speech, and language learning tools.}

\begin{document}
\renewcommand{\thefootnote}{\fnsymbol{footnote}}
\footnotetext[3]{\textbf{Sadded} (\<سَدِيد>): This term conveys the notion of sound judgment, precision, or correctness. Phonetically, it can be rendered in English as \textipa{/s{\ae}"di:d/}.}

\section{Introduction}
\label{sec:introduction}

Diacritization, known as 'tashkīl' in Arabic, plays a crucial role in disambiguating text. It serves as the primary vocalization system for distinguishing between words that share the same consonantal structure but differ in meaning and pronunciation. Table \ref{tab:kalab_represent} illustrates this concept with examples of Arabic words that are visually identical but convey different meanings when diacritized. The importance of diacritization goes beyond mere disambiguation. It is essential to improve various Arabic NLP tasks, including Text-to-Speech (TTS) synthesis \cite{9274427, 9585619}, machine translation \cite{diab2007arabic, zakraoui2021arabic, Fadel_2019}, morphological analysis \cite{9585619, habash2016exploiting}, and Part-of-Speech (POS) tagging \cite{9585619,sabtan2021arabic}. By resolving ambiguities, diacritization significantly improves the accuracy and effectiveness of these downstream tasks.

\begin{table}[htbp]
\centering
\begin{tabular}{lrrrr}
\toprule
\textbf{Arabic word} & \textbf{Pronunciation} & \textbf{Part of speech} & \textbf{English Translation} \\
\midrule
\<قُلِبَ> & \textipa{/kUlIb@/} & Verb & Was turned \\ 
\<قَلَبَ> & \textipa{/k\textturnv l\textturnv b@/} & Verb & Turn \\
\<قَلَّبَ> & \textipa{/k\textturnv ll\textturnv b@/} & Verb & Flip  \\
\<قَلْب> & \textipa{/k\textturnv lb/}  & Noun & Heart \\
\<قُلَّب> & \textipa{/kUll\textturnv b/} & Noun & Volatile \\
\<قُلُب> & \textipa{/kUlUb/} & Noun & Plural of heart \\
\midrule
\end{tabular}
\caption{\small{Comparison of meanings and pronunciations using diacritics. The table illustrates differences in phonetic transcriptions and their associated meanings, highlighting how diacritics affect pronunciation and meaning.}}
\label{tab:kalab_represent}
\end{table}

However, the field of Arabic diacritization faces several challenges:
\begin{itemize}
    \item Data Scarcity: There is a notable lack of labeled data. Modern Arabic writing often omits diacritics for efficiency, as adding them can require multiple keystrokes per character, making the process time-consuming and labor-intensive.
    \item Writing Styles: The Arabic Text is typically categorized into Classical Arabic (CA) and Modern Standard Arabic (MSA). CA refers to historical manuscripts, while MSA represents contemporary writing styles. Although most available diacritization datasets are in Classical Arabic, training diacritization models predominantly on CA data degrades their performance in Modern Standard Arabic. 
    \item Contextual Dependence: Accurate diacritization often requires understanding the full sentence context, a factor sometimes considered in existing models. Table \ref{tab:context_explain} illustrates how Arabic diacritics can change based on the context of subsequent words in a sentence, dramatically altering meaning and grammatical function.
    \item Benchmark Limitations: Current benchmarks often focus exclusively on either CA or MSA, lacking a comprehensive representation of both styles or the diversity of text genres. Additionally, some widely-used benchmarks have been found to contain errors in data splitting and labeling.
\end{itemize}

\begin{table}[ht]
    \centering
    \begin{tabular}{c}
     \includegraphics[
     clip,
     % left bottom right top
     trim=0cm 16.3cm 0cm 2cm, 
     width=1\textwidth]{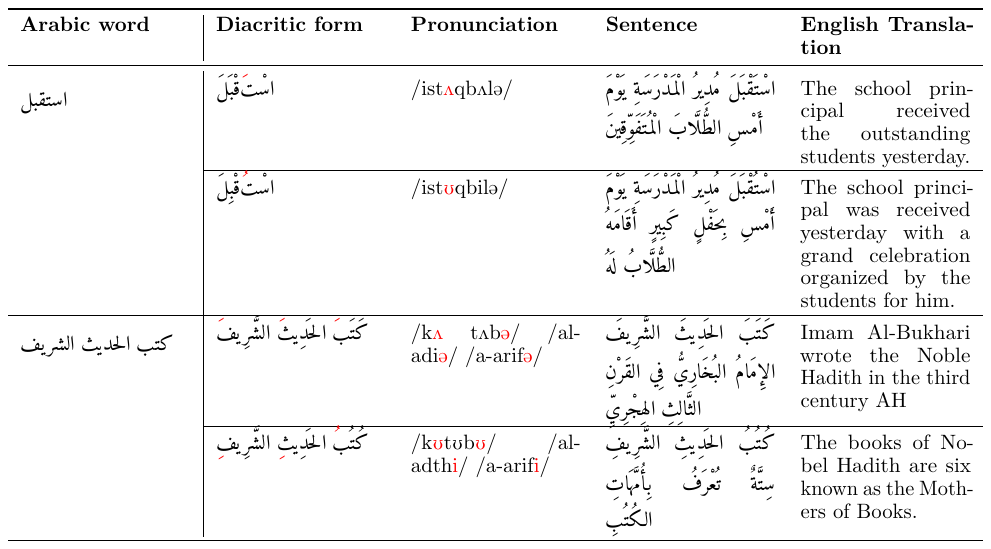}\\
    \end{tabular}
     \caption{\small{illustrates how diacritics in Arabic can change based on the context of subsequent words in a sentence, dramatically altering meaning and grammatical function}}
    \label{tab:context_explain}
\end{table}

In addition to these well-known challenges, there are other less frequently mentioned issues, such as the ambiguity related to punctuation. This problem can significantly affect the correct diacritization of Arabic text, as illustrated in Table \ref{tab:punctuation_ambiguity}, which provides examples of how punctuation can alter the diacritization and meaning of sentences.

\begin{table}[H]
    \centering
    \begin{tabular}{c}
     \includegraphics[
     clip,
     % left bottom right top
     trim=0cm 19.6cm 0cm 2cm, 
     width=1\textwidth]{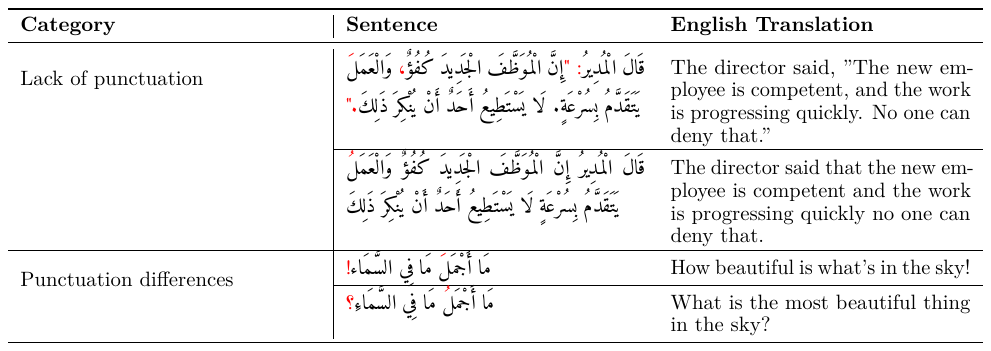}\\
    \end{tabular}
     \caption{\small Punctuation can cause diacritization ambiguities in Arabic by altering sentence structure and grammatical roles, as seen in shifts between case endings and changes in word function (e.g., verb vs. adjective).}
    \label{tab:punctuation_ambiguity}
\end{table}

To address these challenges, we introduce \textbf{Sadeed}, a compact and task-specific Arabic diacritization model fine-tuned on high-quality, noise-free diacritized text. Sadeed is adapted from \textbf{Kuwain}\cite{hennara2025kuwain15barabicslm}, a decoder-only small language model (SLM) pre-trained on diverse Arabic corpora. Optimized for accurate performance on clean, structured input, Sadeed demonstrates competitive results compared to proprietary models. We also propose \textbf{SadeedDiac-25},, a novel open-source benchmark. SadeedDiac-25 incorporates both Classical Arabic and Modern Standard Arabic texts, carefully curated and reviewed by experts to avoid the pitfalls of previous benchmarks. This comprehensive approach aims to provide a more accurate and reliable evaluation framework for Arabic diacritization systems.

In summary, our main contributions are:
\begin{itemize}
    \item \textbf{Sadeed}: a compact decoder-only language model fine-tuned specifically for Arabic diacritization. Despite its small size and limited training resources, Sadeed achieves competitive results against proprietary models and outperforms traditional models trained in-domain.
    \item \textbf{SadeedDiac-25}: We present a novel, comprehensive benchmark for Arabic diacritization, which includes:
    \begin{itemize}
        \item A balanced representation of both Classical Arabic (CA) and Modern Standard Arabic (MSA) texts.
        \item Carefully curated and expert-reviewed data to ensure accuracy and reliability.
    \end{itemize}
    \item \textbf{Error Analysis of existing benchmarks}: We conduct a thorough analysis of existing benchmarks \textit{Abbad and Xiong} \cite{Abbad} and \textit{CATT} \cite{catt}, identifying errors in data splitting, labeling, and content.
    \item \textbf{Diacritization Dataset}: We also release a high-quality, publicly available dataset derived from the \textit{Tashkeela Corpus} \cite{tashkeela}, which has been rigorously filtered and cleaned to ensure consistency and reliability. While it is compatible with the commonly used \textit{Fadel} test set \cite{fadel} for evaluation, our dataset significantly expands the training pool, offering approximately 1 million diacritized examples and  approximately 53 million words. This resource is intended to support the development and benchmarking of Arabic diacritization models at scale.
\end{itemize}
Together, these contributions provide a more robust foundation for advancing research in Arabic diacritization. They also highlight the importance of tailored benchmarks, clean training data, and domain-specific modeling in achieving reliable and reproducible performance in this linguistically complex task. 

\section{Related work}
\label{sec:related}
Previous research has examined a broad array of methodologies for the Arabic diacritization task, evolving from \textit{rule-based} approaches \cite{nelken-shieber-2005-arabic, pasha-etal-2014-madamira, ALNEFAIE2017169} to traditional \textit{machine learning} models \cite{darwish-etal-2017-arabic, abdelali-etal-2016-farasa}  and, more recently, to sophisticated \textit{deep learning} architectures.   In this section, we focus on these advanced methods, categorizing them into two primary classes: \textbf{Recurrent-based} Models and \textbf{Transformer-based} Models, which have proven most effective in addressing the complexities of Arabic diacritization. 

\subsection{Recurrent-based Models}
Recurrent neural networks, particularly LSTM and Bidirectional LSTM (BiLSTM) architectures, have demonstrated significant success in modeling long-range dependencies within sequential data, a crucial aspect of the diacritization task. These models have demonstrated the ability to model the contextual information necessary for accurate diacritization effectively. The following section reviews a range of influential studies that adopt such models:
Several deep learning models based on feed-forward neural networks (FFNNs) and recurrent neural networks (RNNs) that are proposed in \cite{fadel}, integrating techniques such as 100-hot encoding, embeddings, conditional random fields (CRF), and Bayesian Neural Gates (BNG). The models are trained on an adapted version of the Tashkeela corpus, with additional data from the Classical Arabic section of Tashkeela and the Holy Quran. Evaluation is conducted on the original "Fadel Tashkeela" test set as well as datasets from the Al-Shamela Library and the MSA section of Tashkeela.
A sequence-to-sequence model utilizing a BiLSTM-CRF architecture is proposed in \cite{al2020arabic}, adopting a character-level input-output framework for predicting diacritics. The model is trained on four distinct datasets: the King Abdulaziz City for Science and Technology Text-to-Speech (KACST TTS) dataset, the Holy Quran, Sahih Al-Bukhari, and the Penn Arabic Treebank (ATB). Evaluation is performed on a representative subset of this combined corpus.
A hierarchical architecture incorporating cross-level attention mechanisms is introduced in \cite{alkhamissi2020deep}, separating processing at the word and character levels.  This model is trained and evaluated on the "Fadel Tashkeela" dataset.
In \cite{darwish2020arabicdiacriticrecoveryusing}, a feature-rich BiLSTM model is proposed, emphasizing the critical role of integrating linguistic features for diacritization. The model is trained on two distinct corpora: a 9.7 million-token corpus for Modern Standard Arabic (MSA) and a 2.7 million-token corpus for Classical Arabic (CA). The results demonstrate the significant impact of training data size on the model's performance, underscoring the importance of large, high-quality datasets for accurate diacritization.
In \cite{Abbad}, a model with four BiLSTM layers is introduced. The model is trained on a subset of the Abbad training set, with performance evaluation conducted using the first file from the Abbad test set as the benchmark. 
A recent model, 2SDiac \cite{bahar2023hintimprovingarabicdiacritization}, combines bidirectional LSTM (BiLSTM) and self-attention mechanisms with a Guided Learning strategy. This model is trained on the Tashkeela corpus and Arabic Treebanks (ATB) and evaluated on the Tashkeela, ATB, and WikiNews datasets.

These studies collectively underscore the central role of recurrent architectures—particularly BiLSTM-based models—in advancing Arabic diacritization, highlighting their capacity to effectively capture contextual dependencies and generalize across diverse linguistic domains.

\subsection{\textbf{Transformer-based Models} }
Transformer-based Models have emerged as a powerful paradigm in Arabic diacritization, leveraging the strengths of large pre-trained language models and transfer learning techniques. Recent studies have demonstrated the efficacy of these approaches in significantly improving diacritization accuracy. 
In \cite{9274427}, three deep learning models are proposed for Arabic text diacritization: a baseline LSTM model, an encoder-decoder model inspired by Neural Machine Translation (NMT), and an efficient encoder-only model. These models were trained and evaluated on the Tashkeela corpus. The results show that the encoder-only model outperforms the other architectures, highlighting its efficiency and effectiveness in handling the diacritization task.
Furthermore, \cite{skiredj2024arabictextdiacritizationage} introduced a method called PTCAD,  witch is (Pre-FineTuned Token Classification for Arabic Diacritization), which integrates transfer learning with token classification. This approach, trained and validated on the "Abbad Tashkeela" and "Fadel Tashkeela datasets, highlights the increasing trend of applying transfer learning techniques to improve the performance of Arabic diacritization tasks.
In addition, \cite{catt} introduced fine-tuned transformer architectures, both encoder-only and encoder-decoder, using a character-based BERT model. This model employs the Noisy-Student method to enhance performance and is further fine-tuned on the Tashkeela dataset, with evaluations conducted on Wikinews and a newly created CATT dataset.
Lastly, the study \cite{KHARSA2024123416} presents a BERT-based model pre-trained on extensive Arabic corpora and fine-tuned specifically for diacritization. This model is trained and evaluated on the "Abbad Tashkeela" and "Fadel Tashkeela" datasets, Their results underscore the effectiveness of leveraging pre-trained contextual embeddings for improved diacritization accuracy. 

As seen across the reviewed studies, there exists a wide range of training algorithms, architectural choices, and methodological strategies for Arabic diacritization. We have emphasized the training and evaluation datasets employed in each case, as these decisions significantly influence the reported outcomes. By analyzing these aspects, we identify recurring issues—particularly in the use of overlapping data between training and test sets. In Section \ref{subsec:data_analysis}, we delve deeper into these concerns, surveying commonly used benchmarks and highlighting problematic evaluation practices that may compromise the reliability of comparative results.

\section{Diacritization Dataset}
\label{sec:data}
Sadeed training dataset leverages both the Tashkeela corpus \cite{tashkeela} and the Arabic Treebank (ATB-3) \cite{maamouri-etal-2008-diacritic}. Tashkeela is the largest publicly available corpus for the diacritization task, comprising approximately 75 million words, predominantly in classical Arabic. Only about 1.15\% of the text is in Modern Standard Arabic (MSA). In contrast, ATB-3 includes around 300,000 words, primarily in MSA and sourced from news articles.

Several cleaned and preprocessed versions of these datasets exist; however, many are constrained either by limited size—such as the "Fadel Tashkeela" dataset \cite{fadel}—or by flaws in the cleaning and segmentation process. For instance, the "Abbad" version of Tashkeela \cite{Abbad} exhibits sentence truncation that disrupts contextual continuity, a factor critical for achieving accurate diacritization.

To optimize the dataset for this task, we implemented a rigorous preprocessing pipeline:

\textbf{Text Cleaning:}
As noted by Fadel \cite{fadel}, the Tashkeela corpus \cite{tashkeela} suffers from various issues related to text quality and diacritization consistency. To address these challenges, we applied a comprehensive preprocessing pipeline aimed at cleaning the data, correcting diacritization errors, and normalizing the Arabic text. While our process is similar to Fadel's, we aimed to maintain the original text's integrity by preserving non-Arabic characters and other symbols that may occur in Arabic texts, as \textit{Kuwain} is capable of handling such data.

The preprocessing pipeline began by applying the same rigorous cleaning function used in the pretraining of the base model Kuwain \cite{hennara2025kuwain15barabicslm}, followed by additional normalization steps to ensure consistency in diacritization and eliminate common textual errors. Specifically, we unified the diacritization style by omitting the sukūn (absence of a vowel) on elongation letters and the definite article lām when followed by a sun letter. Furthermore, we corrected the diacritization of frequently occurring words—particularly stop words—that typically appear with incomplete or inconsistent vocalization, despite having a single possible diacritization such as \<فِي> and \<عَنْ>.

Finally, we addressed the issue of \textit{iltiqa' assakinayn} (the occurrence of two adjacent consonants without an intervening vowel), which is inconsistently handled across the corpus—some texts apply the phonological rule correctly, while others do not. To resolve this inconsistency, we automatically adjusted the vowelization of the first consonant based on standard Arabic phonological rules.

\textbf{Text Chunking:}
We segmented the corpus into coherent chunks (samples) of 50-60 words each. To achieve this, we employed a hierarchical approach that prioritized splitting and rejoining text chunks using multiple separators (end-of-sentence punctuation marks, line breaks, quotation marks, parentheses, and commas). This method was designed to preserve the syntactical dependencies of the chunk words as much as possible.
By using this hierarchical method, we aimed to avoid splitting the text at punctuation marks that might disrupt the syntactical coherence within each chunk. For example, we prioritized splitting at stronger punctuation marks or natural breaks in the text before resorting to splitting at commas, which often connect closely related clauses or phrases.

\textbf{Dataset filtering:}
The dataset filtering process involves two main steps to optimize the training samples with fully diacritized words. Initially, we excluded examples with more than two words lacking diacritical marks, preserving 93\% of the dataset. Additionally, we removed examples containing three or more words with partial diacritics. After this filtering, only 10.7\% of the sentences in the dataset contain at most two partially diacritized words, ensuring a high standard of diacritic completeness.
To effectively use Fadel \cite{fadel} test set, which is already a split of Tashkeela \cite{tashkeela}, we eliminated overlapping examples with this test set. This process reduced the overlap to only 0.4\% of examples, ensuring that sentences in each sample have at most two overlapping words with Fadel test set.

The resulting dataset comprises 1,042,698 examples totaling approximately 53 million words. It is normalized, contains minimal missing diacritics, and chunked to preserve syntactic and contextual dependencies. This dataset can be used as training data alongside Fadel’s test set without the risk of data leakage, thus providing a solid foundation for models' training and evaluation. The data is made publicly available\footnote[1]{\url{https://huggingface.co/datasets/Misraj/Sadeed_Tashkeela}}.

\section{Benchmark}
\label{sec:benchmark}

Existing benchmarks for Arabic diacritization present notable structural and linguistic limitations that hinder comprehensive evaluation.

First, we observed a clear divide among the benchmarks: some focused exclusively on Classical Arabic (CA), such as Fadel’s \cite{fadel} and Abbad’s \cite{Abbad}, both of which are subsets of the Tashkeela corpus \cite{tashkeela}, while others targeted only Modern Standard Arabic (MSA), such as CATT \cite{catt} and WikiNews \cite{wikinews}. This separation fails to capture the full spectrum of Arabic language usage, limiting the generalizability of models trained on these datasets.

Moreover, several of these corpora exhibit notable flaws. Abbad Tashkeela \cite{Abbad}, for instance, suffers from segmentation issues—discussed earlier—which disrupt contextual coherence. The CATT benchmark \cite{catt} contains numerous spelling and grammatical errors; Section \ref{sec:catt_benchmark_analysis} provides a detailed analysis. As for WikiNews \cite{wikinews}, its primary limitation lies in being derived exclusively from news articles, offering limited domain diversity. Additionally, it contains a large number of named entities, many of which show inconsistent diacritization, particularly in the final letters, often defaulting to sukūn inappropriately.

\begin{figure}[ht]
\centering
\includegraphics[width=0.6\linewidth]{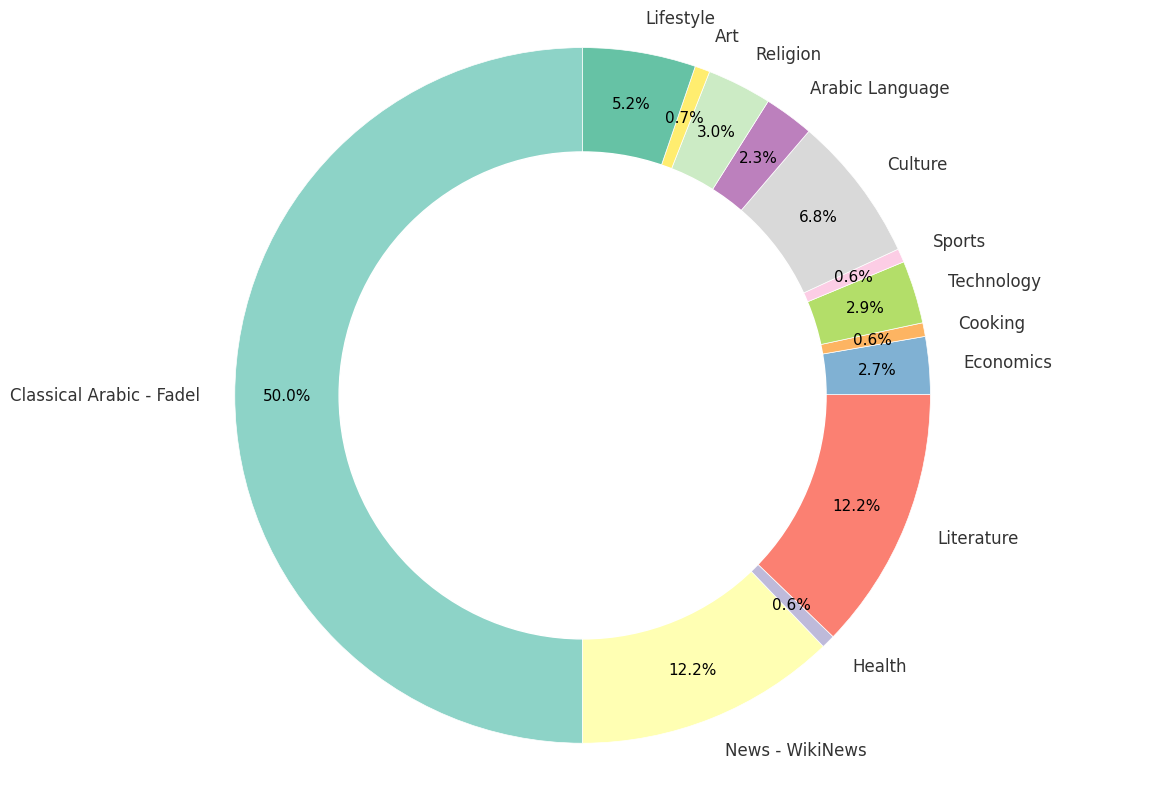}
\caption{\small{Benchmark Category Statistics}}
\label{fig:data_description}
\end{figure}

To address these shortcomings, we introduce \textbf{SadeedDiac-25}—a comprehensive benchmark that unifies both Modern Standard Arabic (MSA) and Classical Arabic (CA) within a single dataset. Our benchmark spans a wide range of topics and writing styles, capturing linguistic diversity across domains such as sports, politics, religion, and culinary arts, as illustrated in Figure~\ref{fig:data_description}.

The development of \textbf{SadeedDiac-25} followed a rigorous and systematic process to ensure high-quality data. As illustrated in Figure~\ref{fig:data_description}, we sourced texts from a variety of domains to achieve broad linguistic and topical coverage. The benchmark comprises 1,200 paragraphs. Of this data, 50\% is in Modern Standard Arabic (MSA)—including 454 paragraphs of originally curated content and 146 paragraphs from WikiNews \cite{wikinews}. These MSA samples range from 40 to 50 words in length. The remaining 50\% represents Classical Arabic (CA), consisting of 600 paragraphs drawn from the Fadel test set \cite{fadel}.
The original content included in the benchmark was curated through the following multi-stage process:
\begin{itemize}
\item \textbf{Data Collection}: Text was sourced from a diverse range of web articles spanning various domains to ensure topic variability and linguistic richness.
    \item \textbf{Initial Diacritization}: The collected text was automatically diacritized using a proprietary large language model. This initial step significantly accelerated the overall process by providing a strong baseline that experts could efficiently refine, rather than starting from unannotated text.
\item \textbf{Expert Review}: A two-stage expert review process was applied to ensure the quality and accuracy of the diacritization:
\begin{itemize}
\item \textit{First Stage}: Two independent experts reviewed the auto-diacritized text, correcting any errors or inconsistencies. 
\item \textit{Second Stage}: Each expert then reviewed the other’s corrections to cross-validate the annotations and resolve any remaining discrepancies or ambiguities.
\end{itemize}
\end{itemize}
This rigorous pipeline ensured that the benchmark content is both linguistically accurate and representative of real-world Arabic usage across various contexts. 

By conducting the diacritization process in-house and incorporating expert review, we ensure that the dataset remains entirely novel to existing models, as they have not been exposed to the diacritized form of our data. This directly addresses the concern raised in the CATT \cite{catt} regarding the potential familiarity of language models with benchmark datasets. Our approach effectively prevents data leakage and results in a genuinely unseen evaluation set, thereby enabling a more reliable and accurate assessment of models' true diacritization capabilities. 
\textbf{SadeedDiac-25} is publicly available\footnote[1]{\url{https://huggingface.co/datasets/Misraj/SadeedDiac-25}} to support further research and development in Arabic diacritization and to facilitate fair and reproducible evaluation of diacritization models.

\section{Model}
\label{sec:method}

\begin{figure}[ht]
    \centering
    \includegraphics[width=0.7\linewidth]{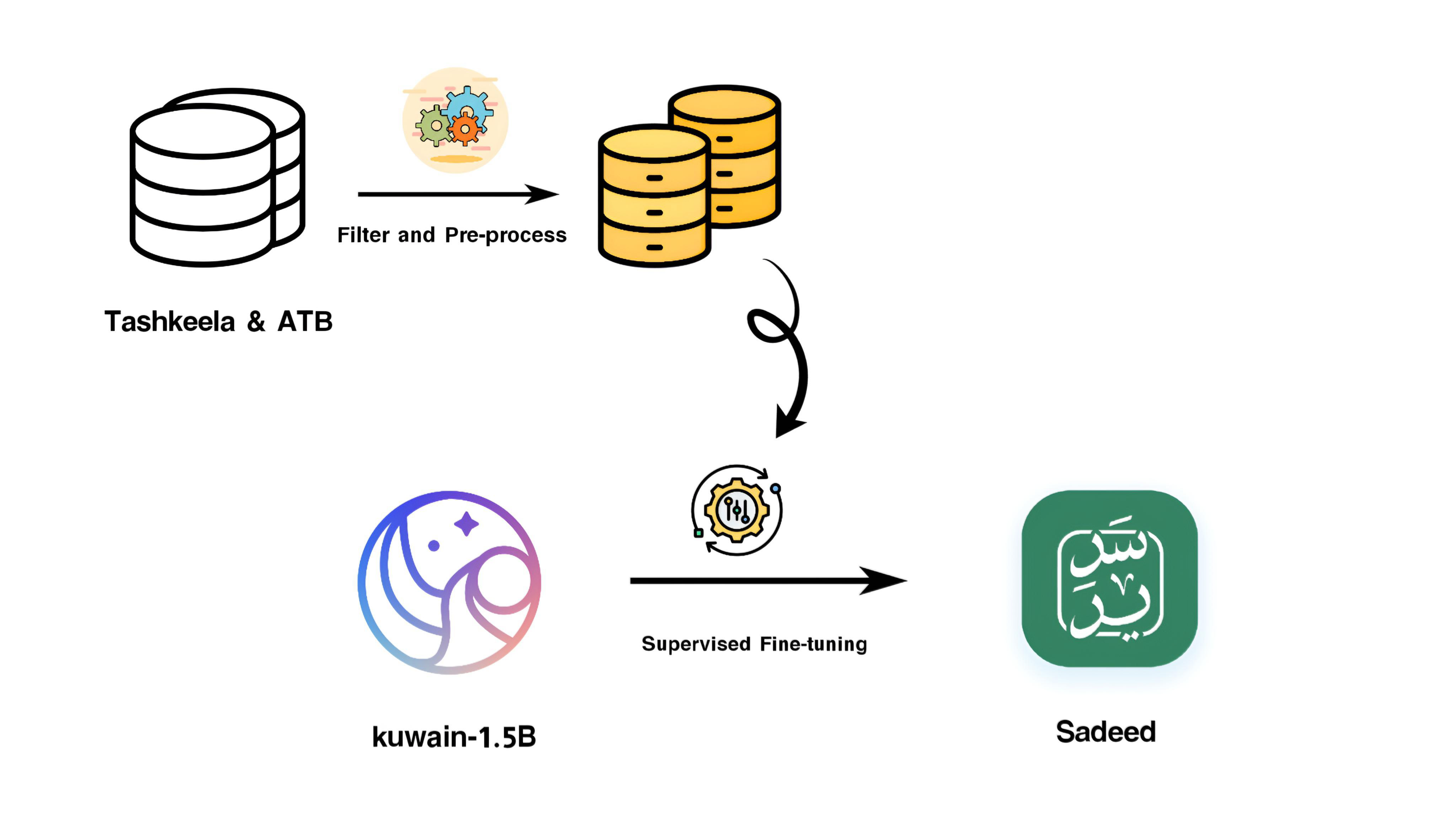}
    \caption{\small{Sadded pipeline, it shows how we get sadeed from our model kuwain-1.5}}
    \label{fig:method}
\end{figure}

\begin{figure}[htbp]
\centering
% First Subfigure
\begin{subfigure}[t]{0.45\textwidth}
\centering
\includegraphics[width=\textwidth]{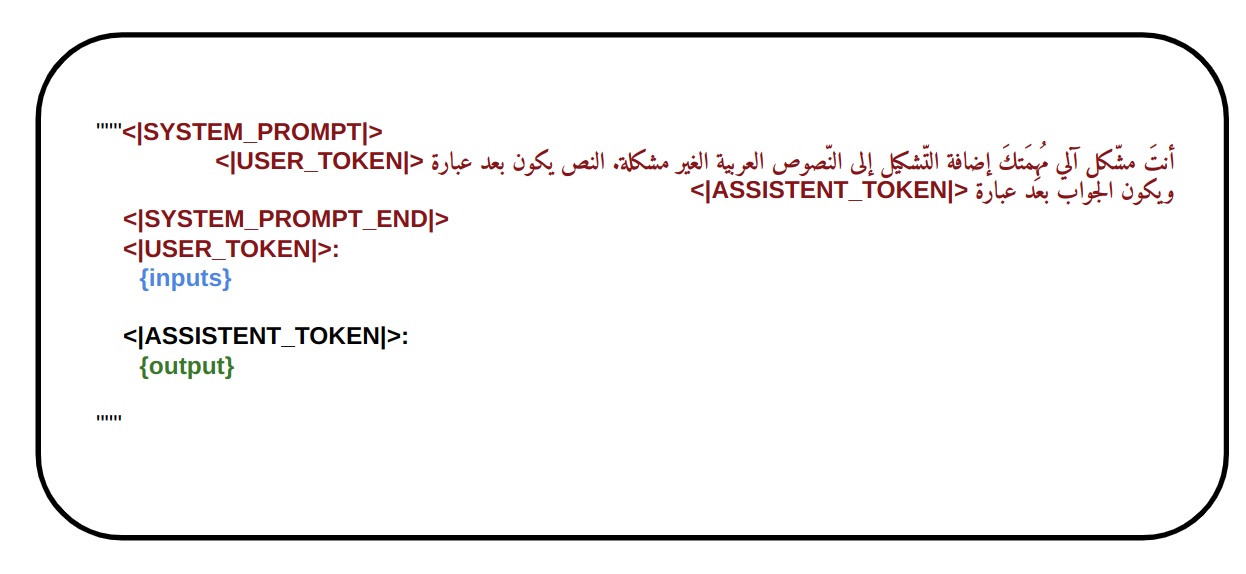}
\caption*{(a)}
\label{fig:image1}
\end{subfigure}
\hfill
% Second Subfigure
\begin{subfigure}[t]{0.45\textwidth}
\centering
\includegraphics[width=\textwidth]{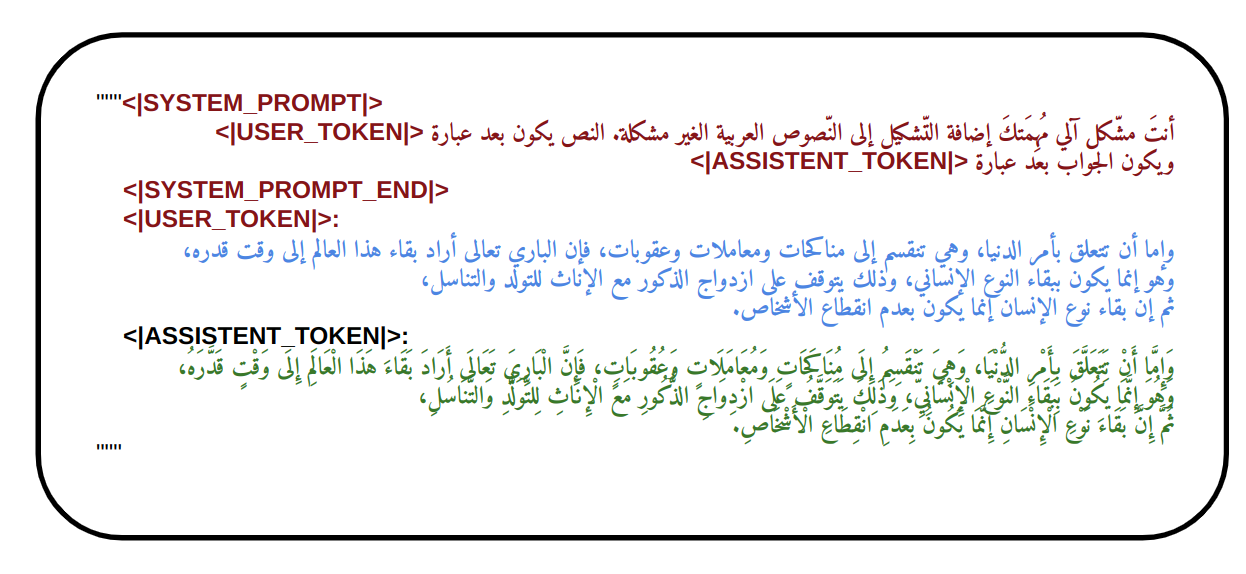}
\caption*{(b)}
\label{fig:image2}
\end{subfigure}
\caption{\small{(a) is The System prompts we use to fine-tune our models. (b) An example from our dataset after applying the template}}
\label{fig:side_by_side}
\end{figure}

\textbf{Sadeed} is a fine-tuned variant of \textbf{Kuwain} \cite{hennara2025kuwain15barabicslm}, a small language model (SLM) specifically tailored for Arabic using an original language injection technique. The fine-tuning process was carefully designed to optimize the model for the task of Arabic diacritization. As part of this process, we reformulated diacritization as a Question-Answering (QA) task, enabling more focused and efficient training by leveraging the model’s generative capabilities in a structured manner.
To prepare the model for fine-tuning, we applied a consistent template transformation across the entire training dataset, as illustrated in Figure \ref{fig:side_by_side}. This step was crucial in adapting the general-purpose \textit{Kuwain} model to the specialized task of diacritization. The complete workflow of our approach is presented in Figure \ref{fig:method}.

This fine-tuned version of \textit{Kuwain}, which we call \textit{Sadeed}, demonstrates the potential of adapting compact models for specific NLP tasks. The resulting model offers a balance between performance and efficiency, making it particularly suitable for Arabic diacritization tasks. For detailed information about the training methodology and hyperparameters, please refer to Appendix \ref{app:training}.

During inference, non-diacritized text is input into the model using the provided template in the \textit{{inputs}} field. The model then generates the corresponding diacritized text for the input.
As a generative model, the output generally matches the input. However, there are instances where the generated text differs from the original. This discrepancy, known as hallucination, occurs because the model generates the most probable tokens based on the given context.
To correct the resulting text, we compare the input with the output. We remove any added words, restore missing words, and replace altered words in the output with their non-diacritized versions. To identify these discrepancies, we use the Needleman-Wunsch \cite{likic2008needleman} algorithm for sequence alignment.
The resulting text is a diacritized version of the original, with some words remaining undiacritized due to being hallucinated.  see Appendix \ref{app:C} for examples of the model's hallucinations and Appendix \ref{app:D} for detailed examples of the model's output over various text types.

\section{Analysis of Existing Benchmarks and Training Practices }
\label{subsec:data_analysis}
Despite recent advances in Arabic diacritization, the field still faces critical challenges related to dataset design, benchmark reliability, and training methodologies. Many existing datasets and benchmarks suffer from issues such as inconsistent annotation, errors, and content overlapping between training and test sets. These issues undermine the validity of the evaluation results and complicate fair comparisons between models. In this section, we examine the prevailing practices in dataset usage, analyze benchmark construction, and identify limitations that could hinder progress in the field. 

\subsection{Overlap Analysis Between Fadel and Abbad Datasets }
\label{subsec:fadel_and_abbad}
To ensure the validity and integrity of model evaluation, it is essential to examine potential overlaps between training and test data. In this subsection, we analyze the degree of content overlap between two widely used diacritized Arabic datasets: Fadel Tashkeela \cite{fadel} and Abbad Tashkeela \cite{Abbad}. Our analysis reveals substantial overlap in both the training and benchmark subsets of these datasets. This overlap calls into question their joint use in training and evaluation pipelines, as it can lead to data leakage, inflated performance results, and a distorted view of a model’s true generalization capabilities. We quantify the overlap and discuss its implications for benchmark design and best practices in training.

Since the Fadel and Abbad samples differ in length, with Abbad examples being generally shorter, we segmented the Fadel samples using punctuation markers, without accounting for context, to match the segmentation strategy used for Abbad. This allowed for a consistent comparison between the two datasets, focusing on the recurrence of contextual structures rather than individual word repetitions. Similarity is calculated as the ratio of the number of words in the overlapping segments to the total number of words in the original sample.
\begin{itemize}
\item \textit{Overlap (Identical Samples)} refers to:
\begin{itemize}
\item \textit{In the Fadel set}: Instances where all segments of a Fadel sample are found in the Abbad set, resulting in a similarity score of 1.0.
\item \textit{In the Abbad set}: Instances where an Abbad sample exactly matches any segment derived from a Fadel sample.
\end{itemize}
\item \textit{Overlap (Samples with Similarity > 0.5)} is calculated only for Fadel samples. It includes cases where more than 50\% of the words in a Fadel sample appear in segments that overlap with the Abbad dataset. This measure does not apply to Abbad samples, as each consists of a single segment that either exactly matches part of a Fadel sample or does not match at all.
\end{itemize}

Table \ref{tab:abbad_fadel_overlap} provides an overview of the overlap between the Abbad (Train + Validation) set and the Fadel test set. It reveals that 6065 samples from the Abbad training set (0.2\%) are identical to those in the Fadel test set. In contrast, 865 samples from the Fadel test set (34.6\%) are fully present in the Abbad training set. Furthermore, 1703 samples from the Fadel test set (68.12\%) exhibit a similarity score greater than 0.5 when compared to the Abbad training set. 

\begin{table}[htbp]
\small
\centering
\begin{tabularx}{\textwidth}{
    >{\raggedright\arraybackslash}X
    >{\raggedright\arraybackslash}X
    >{\raggedright\arraybackslash}X
    >{\raggedright\arraybackslash}X
}
\toprule
\textbf{Dataset} & \textbf{Total Samples} & \textbf{Overlap (Identical Samples)} & \textbf{Overlap (Similarity > 0.5)} \\
\midrule
Abbad Train & 2,940,000 & 6065 (0.2\%) & — \\
Fadel Test & 2,500 & 865 (34.6\%) & 1703 (68.12\%) \\
\midrule
\end{tabularx}
\caption{\small{Comparison of datasets showing the total number of samples, overlap of identical samples (with percentage), and overlap of samples with similarity greater than 0.5 (with percentage).}}
\label{tab:abbad_fadel_overlap}
\end{table}

Table \ref{tab:fadel_abbad_overlap} shows that 657 samples from the Fadel training set (1.25\%) are completely identical to those from the Abbad test set. Similarly, 7045 samples from the Abbad test set (4.69\%) are fully present in the Fadel training set. Furthermore, 1,345 samples from the Fadel training set (2.56\%) exhibit a similarity score greater than 0.5 compared to the Abbad test set.

\begin{table}[htbp]
\small
\centering
\begin{tabularx}{\textwidth}{
    >{\raggedright\arraybackslash}X
    >{\raggedright\arraybackslash}X
    >{\raggedright\arraybackslash}X
    >{\raggedright\arraybackslash}X
}
\toprule
\textbf{Dataset} & \textbf{Total Samples} & \textbf{Overlap (Identical Samples)} & \textbf{Overlap (Similarity > 0.5)} \\
\midrule
Fadel Train & 52,500 & 657 (1.25\%) & 1345 (2.56\%) \\
Abbad Test & 150,015 & 7045 (4.69\%) & — \\
\midrule
\end{tabularx}
\caption{\small{Comparison of datasets showing the total number of samples, overlap of identical samples (with percentage), and overlap of samples with similarity greater than 0.5 (with percentage).}}
\label{tab:fadel_abbad_overlap}
\end{table}

This overlap raises serious methodological concerns for model evaluation in Arabic diacritization tasks. Researchers should be cautious when using these datasets together in a training-evaluation pipeline, as significant content duplication could artificially inflate performance metrics and provide misleading assessments of model generalization. Notably, such overlaps have been observed in the evaluation setup of the PTCAD model \cite{skiredj2024arabictextdiacritizationage}, where the model is trained and evaluated on both Abbad and Fadel datasets. By providing our cleaned version of the Tashkeela dataset, which removes problematic overlaps and enables fair evaluation on the Fadel benchmark, we effectively address these data leakage concerns. 

\subsection{CATT Benchmark Analysis}
\label{sec:catt_benchmark_analysis}
We critically examined the CATT benchmark introduced in \cite{catt}, which presents the Character-based Arabic Tashkeel Transformer (CATT) model and claims state-of-the-art performance on its own curated benchmark. While the proposed CATT model demonstrates strong results, our analysis reveals substantial limitations in both the benchmark dataset and the evaluation methodology, raising concerns about the generalizability and reliability of the reported performance.
One of the most critical issues with the CATT dataset is the complete removal of punctuation marks. Punctuation plays a vital role in defining sentence boundaries and guiding syntactic and semantic interpretation—both of which are essential for accurate diacritization. Without punctuation, the model loses key contextual cues that help disambiguate words and phrases based on sentence structure. This limitation significantly hinders the model’s ability to leverage broader sentence-level context, thereby constraining its effectiveness and potentially inflating its reported performance in controlled settings.

A linguistics expert conducted a detailed review of 30\% of the CATT evaluation data, identifying several distinct categories of diacritization errors (see Appendix \ref{app:B} for a comprehensive tabulation). These errors can be classified as follows:
\begin{itemize}
\item \textbf{Diacritization ambiguity}: Occurs in situations such as short sentences that prevent correct diacritization, often in passive voice constructions.
\item \textbf{Partially diacritized}: Some letters in certain words are not diacritized.
\item \textbf{Erroneous diacritization}: Incorrect diacritization.
\item \textbf{Misplaced diacritics}: Incorrect placement of vowels. This category includes false-diacritization, where vowels are transposed.
\end{itemize}
The limitations identified in CATT benchmark pose significant methodological concerns, potentially resulting in inaccurate evaluation outcomes. By presenting our carefully curated benchmark, developed through a rigorous and systematic process, we effectively address these critical shortcomings, providing researchers with a more reliable and robust evaluation framework. 

\section{Evaluation}
To thoroughly assess the performance of our Sadeed model, we conducted an extensive series of experiments using well-established benchmarks in Arabic diacritization. This section reports the results of these evaluations, offering a detailed comparison between Sadeed and state-of-the-art models across multiple datasets and evaluation metrics. To ensure transparency and reproducibility, we have made the evaluation code \footnote[2]{\url{https://github.com/misraj-ai/Sadeed}} publicly available.
\subsection{Evaluation on Fadel Benchmark}
We first evaluated Sadeed on the widely-used Fadel \cite{fadel} test split to establish a clear comparison with existing state-of-the-art models. Although the Fadel dataset is considered the most reliable benchmark for classical Arabic, it generally lacks consistency in handling cases of adjacent consonants without intervening vowels (iltiqā` as-sākinayn). To address this limitation, we systematically the vowelization of the first consonant based on standard Arabic phonological rules. This refined and phonologically consistent version of the Fadel test set has been made publicly available to facilitate accurate and reliable benchmarking.

Table \ref{tab:fadel_evaluate} presents our evaluation results in terms of Word Error Rate (WER) and Diacritic Error Rate (DER), comparing Sadeed to existing state-of-the-art models. To ensure a fair and transparent comparison, we report performance metrics on both the original and the phonologically corrected versions of the Fadel test set.

\begin{table}[ht]
\centering
   \fontfamily{ptm}\selectfont
\fontsize{9}{12}\selectfont
\begin{tabular}{|c|c|c|c|c|c|c|c|c|}
\toprule
\multirow{3}{*}{{\bf Paper}} & \multicolumn{4}{c|}{{\bf Including No Diacritic}} & \multicolumn{4}{c|}{{\bf Excluding No Diacritic}} \\ \cline{2-9} 
 & \multicolumn{2}{c|}{{\bf w/case ending}} & \multicolumn{2}{c|}{{\bf w/o case ending}} & \multicolumn{2}{c|}{{\bf w/case ending}} & \multicolumn{2}{c|}{{\bf w/o case ending}} \\ \cline{2-9} 
 &  DER & WER & DER & WER & DER & WER & DER & WER \\ \hline
D3 \cite{alkhamissi2020deep}&1.83  & 5.34 & 1.48 & 3.11 & 2.09 & 5.08  & 1.69 & 3.00 \\ \hline
SUKOUN \cite{KHARSA2024123416} & 1.16 &3.34  & 0.96 & 1.96 & 1.23  & 3.03 & 1.00 & 1.77 \\ \hline
PTCAD \cite{skiredj2024arabictextdiacritizationage}& {\bf 1.1} &  4.19 & - & - & - & - & - & - \\ \hline
Fadel \cite{Fadel_2019}&  1.78 &5.38 & 1.39 & 3.04 &2.05 & 5.17 &1.60 &2.96 \\ \hline
{\bf Sadeed- Fadel original} &  1.6814 &4.4914&0.7889& 1.8256&1.5758&4.1469&0.7761& 1.7955 \\
\hline
{\bf Sadeed- Fadel corrected} &1.2386 & {\bf 2.9375} & {\bf 0.7632} & {\bf 1.7376} & {\bf 1.1432} & {\bf 2.6286} &{\bf 0.7518} &{\bf 1.7115}
\\ 

\bottomrule
\end{tabular}
\caption{\small{Performance comparison of Arabic diacritization models on the Fadel dataset in terms of Diacritic Error Rate (DER) and Word Error Rate (WER)}}
\label{tab:fadel_evaluate}
\end{table}

As shown in Table \ref{tab:fadel_evaluate}, \textbf{Sadeed} achieves state-of-the-art (SOTA) performance on the Fadel dataset in terms of Word Error Rate (WER), particularly when the metric excludes non-diacritized characters in the test set. This notable improvement over previous models underscores the robustness of our approach in handling the intricate challenges of Arabic diacritization.We further argue that our Diacritic Error Rate (DER)  also constitutes SOTA performance, particularly when considering other models ware trained on a combination of the Abbad and Fadel datasets—datasets that exhibit significant overlap, as detailed in Section \ref{subsec:fadel_and_abbad}. This overlap raises concerns about inflated performance metrics due to data leakage. For a more comprehensive discussion, refer to Section \ref{sec:disc}.

\subsection{Evaluation on WikiNews Benchmark}
To ensure a fair comparison of the performance of our model, we evaluated Sadeed on WikiNews dataset \cite{wikinews}, which is widely used for assessing diacritization models on Modern Standard Arabic. This evaluation is particularly important, given that a subset of WikiNews data was also incorporated into our Sadeed-Daic-25 benchmark. By including this assessment, we aim to provide a consistent reference point against existing approaches. Table \ref{tab:wiki_evaluate} presents the performance of \textit{Sadeed} model in comparison to previously reported results in this dataset.

\begin{table}[ht]
\centering
   \fontfamily{ptm}\selectfont
\fontsize{9}{12}\selectfont
\begin{tabular}{|c|c|c|c|c|c|c|c|c|}
\toprule
\multirow{3}{*}{{\bf Paper}} & \multicolumn{4}{c|}{{\bf Including No Diacritic}} & \multicolumn{4}{c|}{{\bf Excluding No Diacritic}} \\ \cline{2-9} 
 & \multicolumn{2}{c|}{{\bf w/case ending}} & \multicolumn{2}{c|}{{\bf w/o case ending}} & \multicolumn{2}{c|}{{\bf w/case ending}} & \multicolumn{2}{c|}{{\bf w/o case ending}} \\ \cline{2-9} 
 &  DER & WER & DER & WER & DER & WER & DER & WER \\ \hline

2SDiac \cite{bahar2023hintimprovingarabicdiacritization}&11.6 &33.2 &9.9 &22.7 &13.2 &32.2 &11.1 &22.1 \\ \hline
CATT \cite{catt} & 5.963 &20.060 &3.631 &11.310& -& -& -& - \\ \hline
FRRNN \cite{darwish2020arabicdiacriticrecoveryusing} &{\bf 3.7} &{\bf 6.0} &{\bf 0.9} & {\bf 2.9} & - & - & - & -  \\ \hline
{\bf Sadeed} & 5.2464 & 14.6352 &3.1107  & 8.4400 & {\bf 3.5895} & {\bf 12.2182} & {\bf 1.8500}  & {\bf 5.8926} \\
\bottomrule
\end{tabular}
\caption{\small{Performance comparison of Arabic diacritization models on the WikiNews dataset in terms of Diacritic Error Rate (DER) and Word Error Rate (WER)}}
\label{tab:wiki_evaluate}
\end{table}

The results presented in Table \ref{tab:wiki_evaluate} show that Sadeed achieves competitive performance in the WikiNews dataset \cite{wikinews}, although it does not outperform the model proposed by \cite{darwish2020arabicdiacriticrecoveryusing}, nor match its performance on the Fadel dataset. This discrepancy can probably be attributed to Sadeed's limited exposure to Modern Standard Arabic (MSA) during training. In contrast, other models, such as that of \cite{darwish-etal-2017-arabic}, were trained on datasets that share the same distribution as WikiNews, giving them a natural advantage. These findings suggest that increasing the amount of MSA-specific training data could significantly enhance model performance on MSA benchmarks. However, this remains a challenge because of the scarcity of publicly available diacritized MSA corpora.

\subsection{Evaluation on the SadeedDiac-25 Benchmark }
A key contribution to our work is the introduction of SadeedDiac-25, a new benchmark specifically designed to offer a fresh, diverse, and rigorous evaluation set for Arabic diacritization. This benchmark addresses key limitations in existing datasets by providing a more comprehensive and representative assessment of model performance across different varieties of Arabic.

To evaluate the effectiveness of current models in Arabic diacritization, we conducted an extensive benchmarking study using SadeedDiac-25. The evaluation covers both our model, Sadeed, and leading proprietary large language models, Claude 3.7 Sonnet, GPT-4, and Gemini-Flash 2.0, and prominent open source Arabic models, namely Aya-8B \cite{aya23}, ALLaM-7B-Instruct \cite{allam}, Yehia-7B \cite{yehia2025}, Jais-13B \cite{jais}, Gemma-2-9B\cite{gemma2}, and SILMA-9B-Instruct-v1.0\cite{silma}.

Each model was prompted to diacritize the texts sampled from SadeedDiac-25. Since raw model output often contains hallucinations, such as missing, altered, or inserted words, we applied the Needleman-Wunsch alignment algorithm to automatically correct structural discrepancies while preserving the diacritization generated by the models.

Table \ref{tab:sadeed_diac_evaluate} presents a comparative overview of the performance of the evaluated models on the SadeedDiac-25 benchmark, offering insights into their relative strengths and limitations in this task.
    
\begin{table}[ht]
\centering
\fontfamily{ptm}\selectfont
\fontsize{9}{12}\selectfont
\begin{tabular}{|c|c|c|c|c|c|} \hline
\multirow{2}{*}{\textbf{Model}} & \multicolumn{2}{|c|}{\textbf{With Case Ending}} & \multicolumn{2}{|c|}{\textbf{Without Case Ending}} & \multirow{2}{*}{\textbf{Hallucinations}} \\ \cline{2-5}
& DER & WER & DER & WER & \\ \toprule
\textbf{Claude-3-7-Sonnet-Latest} & \textbf{1.3941} & \textbf{4.6718} & \textbf{0.7693} & \textbf{2.3098} & \textbf{0.821} \\ \hline
\textbf{GPT-4} & 3.8645& 5.2719& 3.8645& 10.9274& 1.0242\\ \hline
\textbf{gemini-flash-2.0} & 3.1926&  7.9942& 2.3783& 5.5044& 1.1713\\ \hline
\textit{\textbf{Sadeed}}&  \textit{7.2915}&\textit{13.7425}& \textit{5.2625}& \textit{9.9245}&\textit{7.1946}\\ \hline
\textbf{aya-23-8B} &  25.6274&47.4908&19.7584& 40.2478&5.7793\\ \hline
\textbf{ALLaM-7B-Instruct} &50.3586& 70.3369& 39.4100& 67.0920&36.5092\\ \hline
\textbf{Yehia-7B}& 50.8801& 70.2323& 39.7677& 67.1520&43.1113\\ \hline 
\textbf{jais-13B}& 78.6820& 99.7541& 60.7271& 99.5702&61.0803\\ \hline 
\textbf{gemma-2-9b}& 78.8560& 99.7928& 60.9188& 99.5895&86.8771\\ \hline 
\textbf{SILMA-9B-Instruct-v1.0}& 78.6567& 99.7367& 60.7106& 99.5586&93.6515\\ \hline
\end{tabular}
\caption{\small{Performance comparison of Arabic diacritization models on the \textbf{SadeedDiac-25} dataset in terms of Diacritic Error Rate (DER) and Word Error Rate (WER)}}
\label{tab:sadeed_diac_evaluate}
\end{table}

The results in Table \ref{tab:sadeed_diac_evaluate} reveal several important trends. \textit{Claude 3.7 Sonnet} consistently achieved the best performance across all evaluation metrics, producing the fewest diacritization errors and exhibiting minimal hallucination, highlighting its superior generalization and robustness on Arabic text. Compared to open source Arabic models, Sadeed demonstrated the strongest performance, outperforming all other open models by a significant margin and achieving results that are competitive with leading proprietary models. However, the primary limitation of Sadeed lies in its hallucination rate: approximately 7.19 points of its total 9.92 WER are attributed to hallucinated words, likely a consequence of its relatively small model size.

It is also worth noting that Sadeed's results on the SadeedDiac-25 benchmark are lower than its performance on the Fadel test set. This discrepancy can be attributed to the more comprehensive nature of SadeedDiac-25, which covers both Modern Standard Arabic (MSA) and Classical Arabic (CA) texts. Since Sadeed was trained on very limited MSA data, its performance tends to decline when faced with MSA-dominant content, highlighting the need for broader training exposure to achieve consistent results across different Arabic varieties.

The broader set of Arabic open-source models exhibited noticeably bad performance. While they generally understand the diacritization task, they have not been explicitly trained or fine-tuned for it, leading to inconsistent and often suboptimal outputs. Among these, Aya-8B stood out as the best-performing open-source model after Sadeed. Although its performance shows higher variance across samples, it produced fewer hallucinations compared to other models, suggesting that Aya-8B could achieve substantial gains if fine-tuned specifically for diacritization.

Overall, these results highlight the critical role of task-specific training in achieving high diacritization accuracy, and reinforce the value of specialized models such as Sadeed for this challenging and linguistically nuanced task.

\section{Discussion \& Limitations}
\label{sec:disc}

\subsection{Dataset Contamination and Its Implications}
The studies by \citet{KHARSA2024123416} and \citet{skiredj2024arabictextdiacritizationage} utilize the Fadel \cite{fadel} and Abbad \cite{Abbad} datasets for model training, but face a significant challenge due to substantial overlap between these datasets, as both derive from the same Tashkeela \cite{tashkeela} source. Our in-depth analysis in Section \ref{subsec:fadel_and_abbad} reveals high contamination between these benchmarks, raising concerns about previously reported results. This overlap suggests potential inflation of performance metrics, as models may inadvertently be tested on data similar to their training sets. This contamination underscores the critical need for careful dataset curation and highlights the possibility of overly optimistic results in previous research, emphasizing the importance of our SadeedDiac-25 benchmark, which provides a truly independent and diverse evaluation resource for Arabic diacritization models.

\subsection{Challenges with Modern Standard Arabic (MSA)}
Our model demonstrates strong overall performance but shows limitations with Modern Standard Arabic (MSA) text due to insufficient MSA training data. While excelling in Classical Arabic diacritization, it underperforms on MSA content in our SadeedDiac-25 benchmark. This performance gap validates our benchmark's effectiveness in identifying improvement areas for current diacritization models.
To address this limitation, we are expanding our dataset with carefully diacritized MSA texts following our benchmark's rigorous methodology. This effort aims to enhance our model's capabilities across the full spectrum of Arabic language variants.

\subsection{Importance of Rigorous Benchmark Creation}
We urge researchers to exercise rigorous standards when creating Arabic diacritization benchmarks. This task requires a deep understanding of Arabic grammatical rules beyond mere linguistic intuition. Our analysis reveals that even established benchmarks may contain errors significantly impacting model evaluation.
We recommend:
\begin{itemize}
    \item Multi-stage review processes involving expert linguists and grammarians
    \item Diverse datasets spanning Classical and Modern Standard Arabic, with varied topics and writing styles
\end{itemize}
Maintaining high standards for benchmark creation will advance the field toward more accurate and reliable Arabic language processing tools.

\subsection{Limitations}
\label{sec:limitations}
Our study encountered several significant limitations that merit attention. The primary challenge lies in model hallucinations, particularly with non-Arabic words. Addressing these issues necessitates the use of advanced techniques, such as constrained decoding, to ensure greater accuracy and coherence in predictions.
Moreover, resources limitations restricted scaling to larger parameters and performing extended training with high-quality data, which potentially limited the model's performance. However, increasing the model's size could improve capabilities but introduce computational and efficiency challenges.
A notable constraint is the limited availability of openly accessible, high-quality diacritized datasets, particularly for Modern Standard Arabic (MSA). This data scarcity poses a significant barrier to developing models that generalize well across different genres and real-world MSA applications.
In summary, these limitations highlight both the computational and linguistic challenges in Arabic diacritization, emphasizing the need for better resources, architectural optimization, and targeted techniques to improve model robustness, fairness, and generalizability in future work.

\bibliography{main}

\begin{thebibliography}{32}
\providecommand{\natexlab}[1]{#1}
\providecommand{\url}[1]{\texttt{#1}}
\expandafter\ifx\csname urlstyle\endcsname\relax
  \providecommand{\doi}[1]{doi: #1}\else
  \providecommand{\doi}{doi: \begingroup \urlstyle{rm}\Url}\fi

\bibitem[Abbad \& Xiong(2020)Abbad and Xiong]{Abbad}
Hamza Abbad and Shengwu Xiong.
\newblock Multi-components system for automatic arabic diacritization.
\newblock In Joemon~M. Jose, Emine Yilmaz, Jo{\~a}o Magalh{\~a}es, Pablo Castells, Nicola Ferro, M{\'a}rio~J. Silva, and Fl{\'a}vio Martins (eds.), \emph{Advances in Information Retrieval}, pp.\  341--355, Cham, 2020. Springer International Publishing.
\newblock ISBN 978-3-030-45439-5.

\bibitem[Abdelali et~al.(2016)Abdelali, Darwish, Durrani, and Mubarak]{abdelali-etal-2016-farasa}
Ahmed Abdelali, Kareem Darwish, Nadir Durrani, and Hamdy Mubarak.
\newblock {F}arasa: A fast and furious segmenter for {A}rabic.
\newblock In John DeNero, Mark Finlayson, and Sravana Reddy (eds.), \emph{Proceedings of the 2016 Conference of the North {A}merican Chapter of the Association for Computational Linguistics: Demonstrations}, pp.\  11--16, San Diego, California, June 2016. Association for Computational Linguistics.
\newblock \doi{10.18653/v1/N16-3003}.
\newblock URL \url{https://aclanthology.org/N16-3003}.

\bibitem[Al-Thubaity et~al.(2020)Al-Thubaity, Alkhalifa, Almuhareb, and Alsanie]{al2020arabic}
Abdulmohsen Al-Thubaity, Atheer Alkhalifa, Abdulrahman Almuhareb, and Waleed Alsanie.
\newblock Arabic diacritization using bidirectional long short-term memory neural networks with conditional random fields.
\newblock \emph{IEEE Access}, 8:\penalty0 154984--154996, 2020.

\bibitem[Alasmary et~al.(2024)Alasmary, Zaafarani, and Ghannam]{catt}
Faris Alasmary, Orjuwan Zaafarani, and Ahmad Ghannam.
\newblock Catt: Character-based arabic tashkeel transformer, 2024.
\newblock URL \url{https://arxiv.org/abs/2407.03236}.

\bibitem[AlKhamissi et~al.(2020)AlKhamissi, ElNokrashy, and Gabr]{alkhamissi2020deep}
Badr AlKhamissi, Muhammad~N ElNokrashy, and Mohamed Gabr.
\newblock Deep diacritization: Efficient hierarchical recurrence for improved arabic diacritization.
\newblock \emph{arXiv preprint arXiv:2011.00538}, 2020.

\bibitem[Almanea(2021)]{9585619}
Manar~M. Almanea.
\newblock Automatic methods and neural networks in arabic texts diacritization: A comprehensive survey.
\newblock \emph{IEEE Access}, 9:\penalty0 145012--145032, 2021.
\newblock \doi{10.1109/ACCESS.2021.3122977}.

\bibitem[Alnefaie \& Azmi(2017)Alnefaie and Azmi]{ALNEFAIE2017169}
Rehab Alnefaie and Aqil~M. Azmi.
\newblock Automatic minimal diacritization of arabic texts.
\newblock \emph{Procedia Computer Science}, 117:\penalty0 169--174, 2017.
\newblock ISSN 1877-0509.
\newblock \doi{https://doi.org/10.1016/j.procs.2017.10.106}.
\newblock URL \url{https://www.sciencedirect.com/science/article/pii/S1877050917321634}.
\newblock Arabic Computational Linguistics.

\bibitem[Aryabumi et~al.(2024)Aryabumi, Dang, Talupuru, Dash, Cairuz, Lin, Venkitesh, Smith, Campos, Tan, et~al.]{aya23}
Viraat Aryabumi, John Dang, Dwarak Talupuru, Saurabh Dash, David Cairuz, Hangyu Lin, Bharat Venkitesh, Madeline Smith, Jon~Ander Campos, Yi~Chern Tan, et~al.
\newblock Aya 23: Open weight releases to further multilingual progress.
\newblock \emph{arXiv preprint arXiv:2405.15032}, 2024.

\bibitem[Bahar et~al.(2023)Bahar, Gangi, Rossenbach, and Zeineldeen]{bahar2023hintimprovingarabicdiacritization}
Parnia Bahar, Mattia~Di Gangi, Nick Rossenbach, and Mohammad Zeineldeen.
\newblock Take the hint: Improving arabic diacritization with partially-diacritized text, 2023.
\newblock URL \url{https://arxiv.org/abs/2306.03557}.

\bibitem[Bari et~al.(2024)Bari, Alnumay, Alzahrani, Alotaibi, Alyahya, AlRashed, Mirza, Alsubaie, Alahmed, Alabduljabbar, et~al.]{allam}
M~Saiful Bari, Yazeed Alnumay, Norah~A Alzahrani, Nouf~M Alotaibi, Hisham~A Alyahya, Sultan AlRashed, Faisal~A Mirza, Shaykhah~Z Alsubaie, Hassan~A Alahmed, Ghadah Alabduljabbar, et~al.
\newblock Allam: Large language models for arabic and english.
\newblock \emph{arXiv preprint arXiv:2407.15390}, 2024.

\bibitem[Darwish et~al.(2017{\natexlab{a}})Darwish, Mubarak, and Abdelali]{darwish-etal-2017-arabic}
Kareem Darwish, Hamdy Mubarak, and Ahmed Abdelali.
\newblock {A}rabic diacritization: Stats, rules, and hacks.
\newblock In Nizar Habash, Mona Diab, Kareem Darwish, Wassim El-Hajj, Hend Al-Khalifa, Houda Bouamor, Nadi Tomeh, Mahmoud El-Haj, and Wajdi Zaghouani (eds.), \emph{Proceedings of the Third {A}rabic Natural Language Processing Workshop}, pp.\  9--17, Valencia, Spain, April 2017{\natexlab{a}}. Association for Computational Linguistics.
\newblock \doi{10.18653/v1/W17-1302}.
\newblock URL \url{https://aclanthology.org/W17-1302}.

\bibitem[Darwish et~al.(2017{\natexlab{b}})Darwish, Mubarak, and Abdelali]{wikinews}
Kareem Darwish, Hamdy Mubarak, and Ahmed Abdelali.
\newblock {A}rabic diacritization: Stats, rules, and hacks.
\newblock In Nizar Habash, Mona Diab, Kareem Darwish, Wassim El-Hajj, Hend Al-Khalifa, Houda Bouamor, Nadi Tomeh, Mahmoud El-Haj, and Wajdi Zaghouani (eds.), \emph{Proceedings of the Third {A}rabic Natural Language Processing Workshop}, pp.\  9--17, Valencia, Spain, April 2017{\natexlab{b}}. Association for Computational Linguistics.
\newblock \doi{10.18653/v1/W17-1302}.
\newblock URL \url{https://aclanthology.org/W17-1302}.

\bibitem[Darwish et~al.(2020)Darwish, Abdelali, Mubarak, and Eldesouki]{darwish2020arabicdiacriticrecoveryusing}
Kareem Darwish, Ahmed Abdelali, Hamdy Mubarak, and Mohamed Eldesouki.
\newblock Arabic diacritic recovery using a feature-rich bilstm model, 2020.
\newblock URL \url{https://arxiv.org/abs/2002.01207}.

\bibitem[Diab et~al.(2007)Diab, Ghoneim, and Habash]{diab2007arabic}
Mona Diab, Mahmoud Ghoneim, and Nizar Habash.
\newblock Arabic diacritization in the context of statistical machine translation.
\newblock In \emph{Proceedings of Machine Translation Summit XI: Papers}, 2007.

\bibitem[Fadel et~al.(2019{\natexlab{a}})Fadel, Tuffaha, Al-Jawarneh, and Al-Ayyoub]{fadel}
Ali Fadel, Ibraheem Tuffaha, Bara' Al-Jawarneh, and Mahmoud Al-Ayyoub.
\newblock Arabic text diacritization using deep neural networks, 2019{\natexlab{a}}.
\newblock URL \url{https://arxiv.org/abs/1905.01965}.

\bibitem[Fadel et~al.(2019{\natexlab{b}})Fadel, Tuffaha, Al-Jawarneh, and Al-Ayyoub]{Fadel_2019}
Ali Fadel, Ibraheem Tuffaha, Bara’ Al-Jawarneh, and Mahmoud Al-Ayyoub.
\newblock Neural arabic text diacritization: State of the art results and a novel approach for machine translation.
\newblock In \emph{Proceedings of the 6th Workshop on Asian Translation}, pp.\  215–225. Association for Computational Linguistics, 2019{\natexlab{b}}.
\newblock \doi{10.18653/v1/d19-5229}.
\newblock URL \url{http://dx.doi.org/10.18653/v1/D19-5229}.

\bibitem[Habash et~al.(2016)Habash, Shahrour, and Al~Khalil]{habash2016exploiting}
Nizar Habash, Anas Shahrour, and Muhamed Al~Khalil.
\newblock Exploiting arabic diacritization for high quality automatic annotation.
\newblock In \emph{Proceedings of the Tenth International Conference on Language Resources and Evaluation (LREC'16)}, pp.\  4298--4304, 2016.

\bibitem[Hennara et~al.(2025)Hennara, Chrouf, Hamed, Aldallal, Hadid, and AlModhayan]{hennara2025kuwain15barabicslm}
Khalil Hennara, Sara Chrouf, Mohamed~Motaism Hamed, Zeina Aldallal, Omar Hadid, and Safwan AlModhayan.
\newblock Kuwain 1.5 b: An arabic slm via language injection.
\newblock \emph{arXiv preprint arXiv:2504.15120}, 2025.

\bibitem[Kharsa et~al.(2024)Kharsa, Elnagar, and Yagi]{KHARSA2024123416}
Ruba Kharsa, Ashraf Elnagar, and Sane Yagi.
\newblock Bert-based arabic diacritization: A state-of-the-art approach for improving text accuracy and pronunciation.
\newblock \emph{Expert Systems with Applications}, 248:\penalty0 123416, 2024.
\newblock ISSN 0957-4174.
\newblock \doi{https://doi.org/10.1016/j.eswa.2024.123416}.
\newblock URL \url{https://www.sciencedirect.com/science/article/pii/S0957417424002811}.

\bibitem[Likic(2008)]{likic2008needleman}
Vladimir Likic.
\newblock The needleman-wunsch algorithm for sequence alignment.
\newblock \emph{Lecture given at the 7th Melbourne Bioinformatics Course, Bi021 Molecular Science and Biotechnology Institute, University of Melbourne}, pp.\  1--46, 2008.

\bibitem[Maamouri et~al.(2008)Maamouri, Kulick, and Bies]{maamouri-etal-2008-diacritic}
Mohamed Maamouri, Seth Kulick, and Ann Bies.
\newblock Diacritic annotation in the {A}rabic treebank and its impact on parser evaluation.
\newblock In Nicoletta Calzolari, Khalid Choukri, Bente Maegaard, Joseph Mariani, Jan Odijk, Stelios Piperidis, and Daniel Tapias (eds.), \emph{Proceedings of the Sixth International Conference on Language Resources and Evaluation ({LREC}'08)}, Marrakech, Morocco, May 2008. European Language Resources Association (ELRA).
\newblock URL \url{http://www.lrec-conf.org/proceedings/lrec2008/pdf/706_paper.pdf}.

\bibitem[Madhfar \& Qamar(2021)Madhfar and Qamar]{9274427}
Mokthar Ali~Hasan Madhfar and Ali~Mustafa Qamar.
\newblock Effective deep learning models for automatic diacritization of arabic text.
\newblock \emph{IEEE Access}, 9:\penalty0 273--288, 2021.
\newblock \doi{10.1109/ACCESS.2020.3041676}.

\bibitem[Navid-AI(2025)]{yehia2025}
Navid-AI.
\newblock Yehia 7b preview.
\newblock \url{https://huggingface.co/Navid-AI/Yehia-7B-preview}, 2025.

\bibitem[Nelken \& Shieber(2005)Nelken and Shieber]{nelken-shieber-2005-arabic}
Rani Nelken and Stuart~M. Shieber.
\newblock {A}rabic diacritization using weighted finite-state transducers.
\newblock In Kareem Darwish, Mona Diab, and Nizar Habash (eds.), \emph{Proceedings of the {ACL} Workshop on Computational Approaches to {S}emitic Languages}, pp.\  79--86, Ann Arbor, Michigan, June 2005. Association for Computational Linguistics.
\newblock URL \url{https://aclanthology.org/W05-0711}.

\bibitem[Pasha et~al.(2014)Pasha, Al-Badrashiny, Diab, El~Kholy, Eskander, Habash, Pooleery, Rambow, and Roth]{pasha-etal-2014-madamira}
Arfath Pasha, Mohamed Al-Badrashiny, Mona Diab, Ahmed El~Kholy, Ramy Eskander, Nizar Habash, Manoj Pooleery, Owen Rambow, and Ryan Roth.
\newblock {MADAMIRA}: A fast, comprehensive tool for morphological analysis and disambiguation of {A}rabic.
\newblock In Nicoletta Calzolari, Khalid Choukri, Thierry Declerck, Hrafn Loftsson, Bente Maegaard, Joseph Mariani, Asuncion Moreno, Jan Odijk, and Stelios Piperidis (eds.), \emph{Proceedings of the Ninth International Conference on Language Resources and Evaluation ({LREC}'14)}, pp.\  1094--1101, Reykjavik, Iceland, May 2014. European Language Resources Association (ELRA).
\newblock URL \url{http://www.lrec-conf.org/proceedings/lrec2014/pdf/593_Paper.pdf}.

\bibitem[Sabtan(2021)]{sabtan2021arabic}
Yasser Sabtan.
\newblock Arabic part-of-speech tagging using a combined rule-based and data-driven approach.
\newblock \emph{Digital Scholarship in the Humanities}, 36\penalty0 (3):\penalty0 719--735, 2021.

\bibitem[Sengupta et~al.(2023)Sengupta, Sahu, Jia, Katipomu, Li, Koto, Marshall, Gosal, Liu, Chen, Afzal, Kamboj, Pandit, Pal, Pradhan, Mujahid, Baali, Han, Bsharat, Aji, Shen, Liu, Vassilieva, Hestness, Hock, Feldman, Lee, Jackson, Ren, Nakov, Baldwin, and Xing]{jais}
Neha Sengupta, Sunil~Kumar Sahu, Bokang Jia, Satheesh Katipomu, Haonan Li, Fajri Koto, William Marshall, Gurpreet Gosal, Cynthia Liu, Zhiming Chen, Osama~Mohammed Afzal, Samta Kamboj, Onkar Pandit, Rahul Pal, Lalit Pradhan, Zain~Muhammad Mujahid, Massa Baali, Xudong Han, Sondos~Mahmoud Bsharat, Alham~Fikri Aji, Zhiqiang Shen, Zhengzhong Liu, Natalia Vassilieva, Joel Hestness, Andy Hock, Andrew Feldman, Jonathan Lee, Andrew Jackson, Hector~Xuguang Ren, Preslav Nakov, Timothy Baldwin, and Eric Xing.
\newblock Jais and jais-chat: Arabic-centric foundation and instruction-tuned open generative large language models, 2023.
\newblock URL \url{https://arxiv.org/abs/2308.16149}.

\bibitem[Skiredj \& Berrada(2024)Skiredj and Berrada]{skiredj2024arabictextdiacritizationage}
Abderrahman Skiredj and Ismail Berrada.
\newblock Arabic text diacritization in the age of transfer learning: Token classification is all you need, 2024.
\newblock URL \url{https://arxiv.org/abs/2401.04848}.

\bibitem[Team et~al.(2024)Team, Riviere, Pathak, Sessa, Hardin, Bhupatiraju, Hussenot, Mesnard, Shahriari, Ram{\'e}, et~al.]{gemma2}
Gemma Team, Morgane Riviere, Shreya Pathak, Pier~Giuseppe Sessa, Cassidy Hardin, Surya Bhupatiraju, L{\'e}onard Hussenot, Thomas Mesnard, Bobak Shahriari, Alexandre Ram{\'e}, et~al.
\newblock Gemma 2: Improving open language models at a practical size.
\newblock \emph{arXiv preprint arXiv:2408.00118}, 2024.

\bibitem[Team(2024)]{silma}
Silma Team.
\newblock Silma.
\newblock 2024.
\newblock URL \url{https://www.silma.ai}.

\bibitem[Zakraoui et~al.(2021)Zakraoui, Saleh, Al-Maadeed, and Alja’am]{zakraoui2021arabic}
Jezia Zakraoui, Moutaz Saleh, Somaya Al-Maadeed, and Jihad~Mohamed Alja’am.
\newblock Arabic machine translation: A survey with challenges and future directions.
\newblock \emph{IEEE Access}, 9:\penalty0 161445--161468, 2021.

\bibitem[Zerrouki \& Balla(2017)Zerrouki and Balla]{tashkeela}
Taha Zerrouki and Amar Balla.
\newblock Tashkeela: Novel corpus of arabic vocalized texts, data for auto-diacritization systems.
\newblock \emph{Data in Brief}, 11, 02 2017.
\newblock \doi{10.1016/j.dib.2017.01.011}.

\end{thebibliography}
\clearpage
\newpage
\appendix
\section*{Appendix}
\addcontentsline{toc}{section}{Appendix}
\section{Training Details}
\label{app:training}

The fine-tuning process for \textit{Sadeed} employed the standard next-token prediction methodology, with the system prompt and embedding tokens masked. Given the compact nature of the \textit{Kuwain} model, we designed a concise yet effective fine-tuning regimen. Table \ref{tab:training_params} presents the detailed hyperparameters used in our training process.

\begin{table}[h]
\small
\centering
\renewcommand{\arraystretch}{1.2}
\begin{tabular}{p{4.5cm}p{4.5cm}}
\hline
\rowcolor[HTML]{F2F2F2} 
\textbf{Parameter} & \textbf{Value} \\
\hline
\rowcolor[HTML]{F9F9F9}
Training Epochs & 3 \\
Learning Rate Schedule & Cosine decay \\
\rowcolor[HTML]{F9F9F9}
Learning Rate & 5e-6 \\
Batch Size & 1024 \\
\rowcolor[HTML]{F9F9F9}
Weight Decay & 0.01 \\
Warm-up Steps & 30 \\
\rowcolor[HTML]{F9F9F9}
Optimizer & AdamW \\
Max Sequence Length & 512 \\
\hline
\end{tabular}
\caption{\small{Training Hyperparameters for \textit{Sadeed} Model}}
\label{tab:training_params}
\end{table}

The model was trained on 8 A100 GPUs. During fine-tuning, we monitored the validation loss to prevent overfitting and implemented early stopping with a patience of 3 evaluation steps. The best checkpoint was selected based on the lowest validation loss achieved during training.
\newpage
\section{Model Hallucinations Examples}\label{app:C}
In this section, we highlight examples of model hallucination observed during the initial diacritization process. Table \ref{tab:hauucinations} presents side-by-side comparisons between the raw model outputs and the corrected versions obtained through the Needleman-Wunsch alignment algorithm.  

\begin{table}[H]
    \centering
    \begin{tabular}{c}
     \includegraphics[
     clip,
     % left bottom right top
     trim=0cm 17.3cm 0cm 2cm, 
     width=\textwidth]{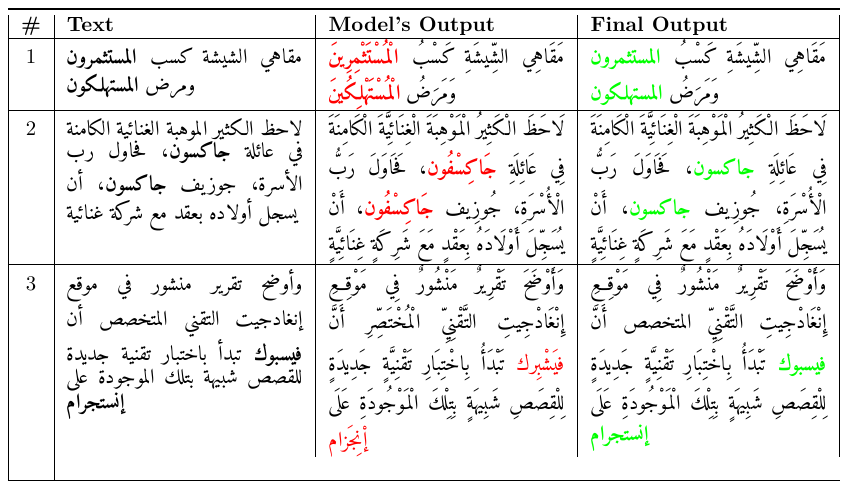}\\
    \end{tabular}
     \caption{\small{Examples of hallucination correction in Arabic text.
     In Sentence 1, the grammatical case ending is corrected to align with the change in word type caused by the hallucinated word. In Sentences 2 and 3, the hallucinated words were foreign terms.}}
    \label{tab:hauucinations}
\end{table}

\newpage
\section{Examples from Sadeed}\label{app:D}
This section provides multiple examples of Sadeed model output. Table \ref{tab:sadeed_sample} shows the results without errors. 
\begin{table}[H]
    \centering
    \begin{tabular}{c}
     \includegraphics[
     clip,
     % left bottom right top
     trim=0cm 7.3cm 0cm 2cm, 
     width=\textwidth]{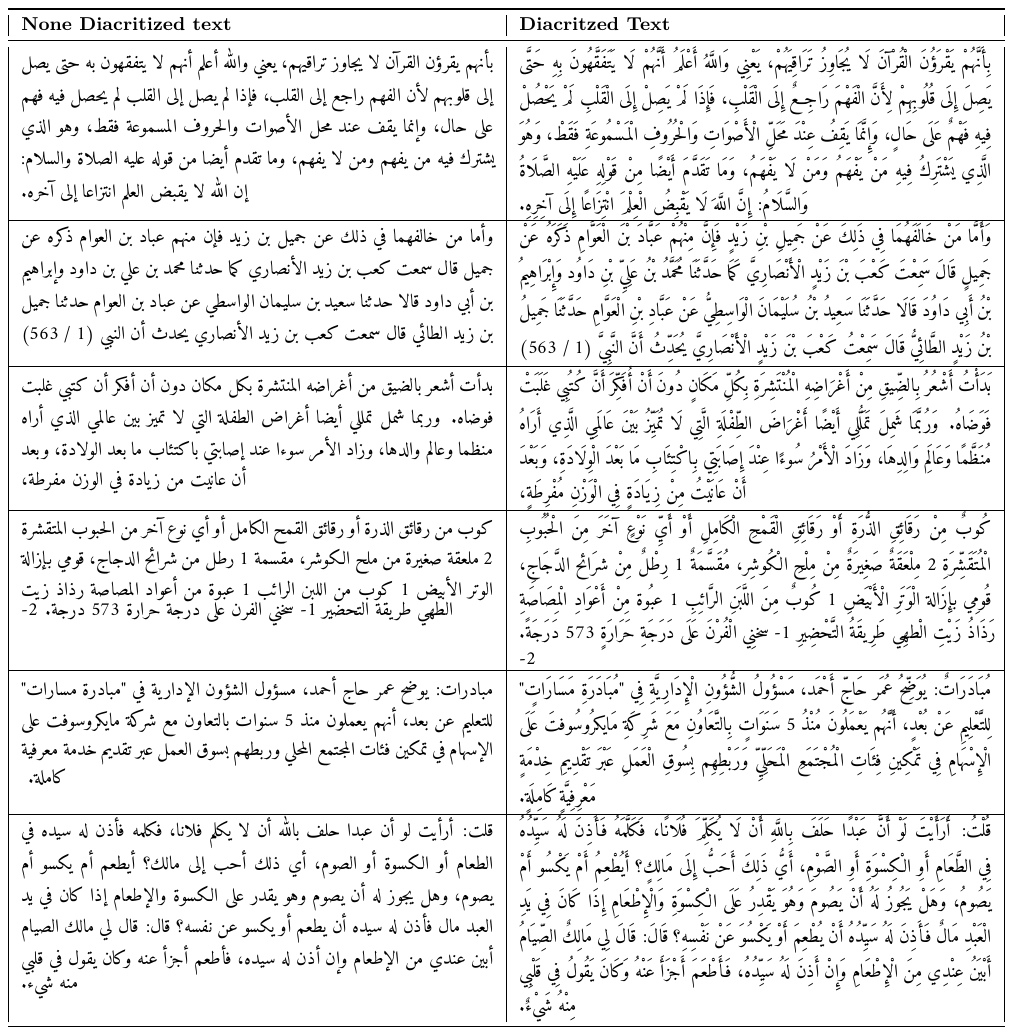}\\
    \end{tabular}
     \caption{\small{Output samples from Sadeed}}
    \label{tab:sadeed_sample}
\end{table}

\newpage
\section{CATT Benchmark Examples}\label{app:B}
Table \ref{tab:catt_errors} illustrates some of errors found in the CATT benchmark, showing examples of incorrect diacritizations, their corrections, and the corresponding error classes.

\begin{table}[H]
    \centering
    \begin{tabular}{c}
     \includegraphics[
     clip,
     % left bottom right top
     trim=0cm 2.3cm 0cm 2cm, 
     width=\textwidth]{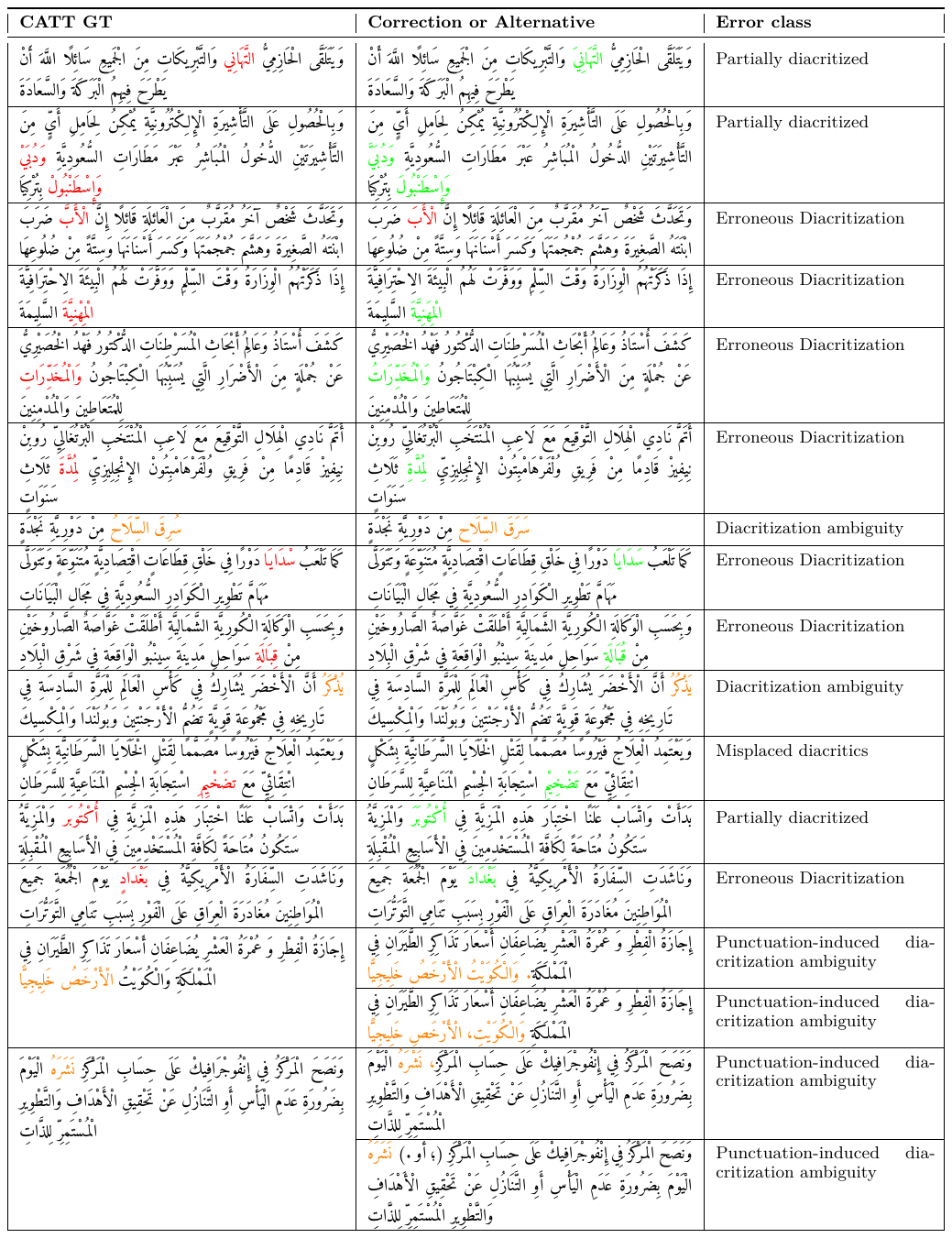}\\
    \end{tabular}
     \caption{\small{In the table provided, the red color indicates the presence of a diacritical error in the word, while the green color denotes the corrected diacritical marking. The orange color represents cases where multiple valid forms are allowed depending on the context or according to different linguistic schools.}}
    \label{tab:catt_errors}
\end{table}

These mistakes range from minor diacritization inconsistencies to more significant issues in data splitting and labeling. For instance, some words are incorrectly diacritized, while others show inconsistent diacritization patterns across similar contexts. Moreover, there are cases where the benchmark's ground truth differs from the correct diacritization according to Arabic linguistic rules. These errors can lead to inaccurate assessments of model performance and potentially misleading conclusions about the state of Arabic diacritization technology.

\setcounter{table}{0}
\setcounter{figure}{0}
\renewcommand{\thetable}{A\arabic{table}}
\renewcommand{\thefigure}{A\arabic{figure}}

\end{document}